%% file: main.tex
\documentclass[conference]{IEEEtran}
\IEEEoverridecommandlockouts
\usepackage{amsmath,amssymb,amsfonts}
\usepackage{algorithmic}
\usepackage{booktabs}
\usepackage{cite}
\usepackage{graphicx}
\usepackage{hyperref}
\usepackage{siunitx}
\usepackage{textcomp}
\usepackage{xcolor}
\def\BibTeX{{\rm B\kern-.05em{\sc i\kern-.025em b}\kern-.08em
    T\kern-.1667em\lower.7ex\hbox{E}\kern-.125emX}}
\begin{document}
\renewcommand{\citedash}{--}
\renewcommand{\subsectionautorefname}{Section}
\renewcommand{\subsubsectionautorefname}{Section}

\title{TS-MoCo: Time-Series Momentum Contrast\\for Self-Supervised Physiological\\Representation Learning}

\author{
    \IEEEauthorblockN{Philipp Hallgarten\textsuperscript{*,1,2},
    David Bethge\textsuperscript{3},
    Ozan Özdenizci\textsuperscript{4,5},
    Tobias Grosse-Puppendahl\textsuperscript{1},
    Enkelejda Kasneci\textsuperscript{2}}
    \IEEEauthorblockA{\textsuperscript{1}Dr. Ing. h.c. F. Porsche AG, Stuttgart, Germany}
    \IEEEauthorblockA{\textsuperscript{2}TU Munich, Germany}
    \IEEEauthorblockA{\textsuperscript{3}LMU Munich, Germany}
    \IEEEauthorblockA{\textsuperscript{4}Institute of Theoretical Computer Science, TU Graz, Austria}%
    \IEEEauthorblockA{\textsuperscript{5}TU Graz - SAL Dependable Embedded Systems Lab, Silicon Austria Labs, Graz, Austria}%
    \thanks{\textsuperscript{*} Corresponding Author: {\tt\small philipp.hallgarten@tum.de}}%
}

\maketitle

\begin{abstract}
Limited availability of labeled physiological data often prohibits the use of powerful supervised deep learning models in the biomedical machine intelligence domain.
We approach this problem and propose a novel encoding framework that relies on self-supervised learning with momentum contrast to learn representations from multivariate time-series of various physiological domains without needing labels.
Our model uses a transformer architecture that can be easily adapted to classification problems by optimizing a linear output classification layer.
We experimentally evaluate our framework using two publicly available physiological datasets from different domains, i.e., human activity recognition from embedded inertial sensory and emotion recognition from electroencephalography.
We show that our self-supervised learning approach can indeed learn discriminative features which can be exploited in downstream classification tasks.
Our work enables the development of domain-agnostic intelligent systems that can effectively analyze multivariate time-series data from physiological domains.
\end{abstract}


\input{sections/01_Introduction.tex}
\input{sections/02_methods.tex}
\input{sections/03_experimental_study.tex}
\input{sections/04_results.tex}
\input{sections/05_discussion.tex}
\section*{Acknowledgement}
This research was partly funded by the Deutsche Forschungsgemeinschaft (DFG, German Research Foundation) in TRR161 (Quantitative methods for visual computing, Project ID 251654672) in Project C06.

\bibliographystyle{IEEEtran}
\bibliography{mybib}

\end{document}

%% file: sections/01_Introduction.tex
\section{INTRODUCTION}

A major challenge in developing high-performing machine learning algorithms in the biomedical domain is the limited availability of labeled data, whereas state-of-the-art deep learning models require more labeled data to generalize better.
This poses a significant problem if such models shall be leveraged to automatize tasks in this domain.
Accordingly, in this paper, we focus on the following:
``\textit{Is it possible to learn physiological domain agnostic deep learning architectures?}" and ``\textit{Can we train such an architecture in a self-supervised manner without the need for labeled data?}".

\begin{figure}[t]
    \centering
 \includegraphics[width=.95\columnwidth]{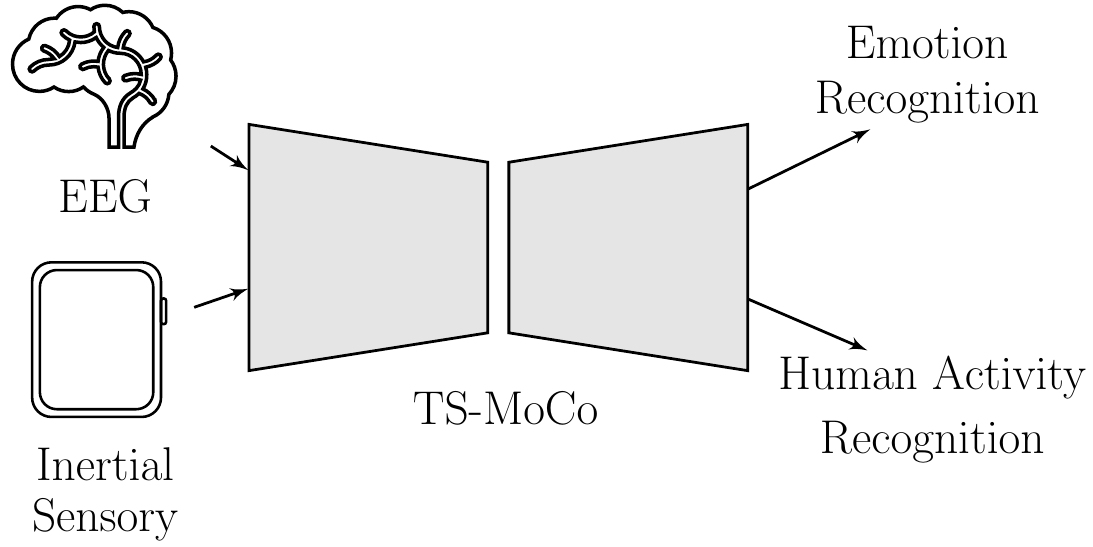}
    \caption{Proposed \textit{TS-MoCo} framework enables optimizing a physiological signal-agnostic encoder architecture for e.g., emotion recognition from EEG or human activity recognition from inertial sensory.}
    \label{fig:visual-abstract}
\end{figure}

One promising avenue in this research domain focuses on deep representation learning~\cite{eldele2021time,han2021universal,ozdenizci2021stochastic,bethge2022eeg2vec}. 
These models learn to extract generalizable features, omitting the need for domain-specific hand-designed features.
We particularly focus on self-supervised learning (SSL) frameworks, a machine learning paradigm where the model is trained without labeled data, \textit{i.e.}, the supervision signal necessary for training is generated on-the-fly from the data itself. 
We investigate the architectural design space in depth and develop a novel SSL framework to learn generalized features across \textit{multiple} physiological signals.
Our proposed deep SSL framework, annotated as \textit{TS-MoCo}, optimizes a general feature encoder agnostic to the physiological domain.
We introduce a two-fold contrastive loss function for optimization based on time-series momentum contrast~\cite{he2020momentum}.
The feature encoder of TS-MoCo predicts physiological labels using only one dense layer that must be fitted on the labeled data on top of the learned embedding space.
In contrast to prior research that focuses on single, specialized domains, we evaluate our framework in two physiological data domains, \textit{i.e.}, emotion classification, and human activity recognition. 
Overall, we address the following questions:
\begin{itemize}
\setlength\itemsep{.6em}
\item \textit{How can we learn a generalizable feature encoder architecture and learning algorithm for diverse domains of physiological signal recordings?}
\item \textit{What is the technical design space for such a deep signal processing framework?}
\item \textit{How are the domains affecting the encoding architecture?}
\end{itemize}

Our framework thereby focuses on learning a signal domain agnostic feature encoder. 
We propose and explore self-supervised learning of general feature encoders as an alternative approach for medical deep learning researchers that enables use in downstream tasks with few labeled data.


\section{RELATED WORK}

\subsection{Deep Learning for Physiological Signal Processing}

In recent years various works explored deep learning based signal processing pipelines for physiological time-series data recordings. 
In~\cite{panagiotou2022comparative}, different autoencoder architectures for encoding physiological signals recorded through smartwatches are compared.
\cite{fix2022transfer} implements a transfer learning approach to train a deep learning model for human activity recognition from radar databases.
Another set of works on physiological signal processing focuses on electroencephalography (EEG) signals.
Cura et al.~\cite{cura2022epileptic} use a deep convolutional neural network, to detect epilepsy from patients EEG recordings, and Geoffroy et al.~\cite{Geoffroy2022drowsiness} detect drowsiness from EEG and electrocardigrams (ECG) with a deep learning architecture.

\subsection{Self-Supervised Learning for Physiological Domains}\label{subsec:related-work-phyisological-domain}

Self-supervised models have been previously explored for various physiological domains since physiological data monitoring is becoming ubiquitous, resulting in large amounts of unlabeled data.
In~\cite{jain2022collossl}, the authors present an SSL framework trained with a contrastive loss that learns representations for human activity recordings of inertial sensory data from multiple devices.
The authors of~\cite{zhang2022ganser} demonstrate that one can successfully improve emotion classification from electroencephalography (EEG) signals by training a generative adversarial network to synthesize data using SSL.
Recently \cite{eldele2021time} proposed a vanilla SSL framework, \textit{TS-TCC}, trained with a combination of temporal and contextual contrastive loss that achieved significant results across several domains.

\subsection{Self-Supervised Learning with Momentum Contrast}
In SSL with momentum contrast, two models of the same architecture are used: a student encoder and a teacher (or momentum) encoder.
The student encoder is trained via loss backpropagation, while the teacher model parameters are set as an exponentially moving average of the student model parameters.
Seminal work in \cite{he2020momentum} proposes \textit{MoCo}, a framework that is trained with momentum contrast. The momentum encoder is hereby used to build a representation dictionary on-the-fly, which is then matched to the representation output by the student encoder network in a contrastive loss. \textit{MoCo} achieves superior performance than supervised baselines on multiple vision tasks.
In the \textit{BYOL} framework by \cite{grill2020bootstrap}, both the momentum encoder and the student encoder predict representations where different input augmentations are applied. \textit{BYOL} achieves comparable results to state-of-the-art baselines in vision tasks while neglecting the need for negative samples.
Finally, \cite{caron2021emerging} proposes \textit{DINO} as an SSL framework trained with momentum contrast without negative samples using vision transformers~\cite{dosovitskiy2020image} as a backbone and evaluated on vision tasks.

Momentum contrast reduces computational demand in comparison to conventional contrastive methods by neglecting negative samples, and exhibits a promising potential.
However, to date, this SSL mechanism is not explored in depth for analyzing biomedical data. 
In our work, we tackle this research gap by proposing a framework based on momentum contrast and evaluating it on different physiological domains.



%




%% file: sections/02_methods.tex
\begin{figure*}[t]
    \centering
    \includegraphics[width=\textwidth]{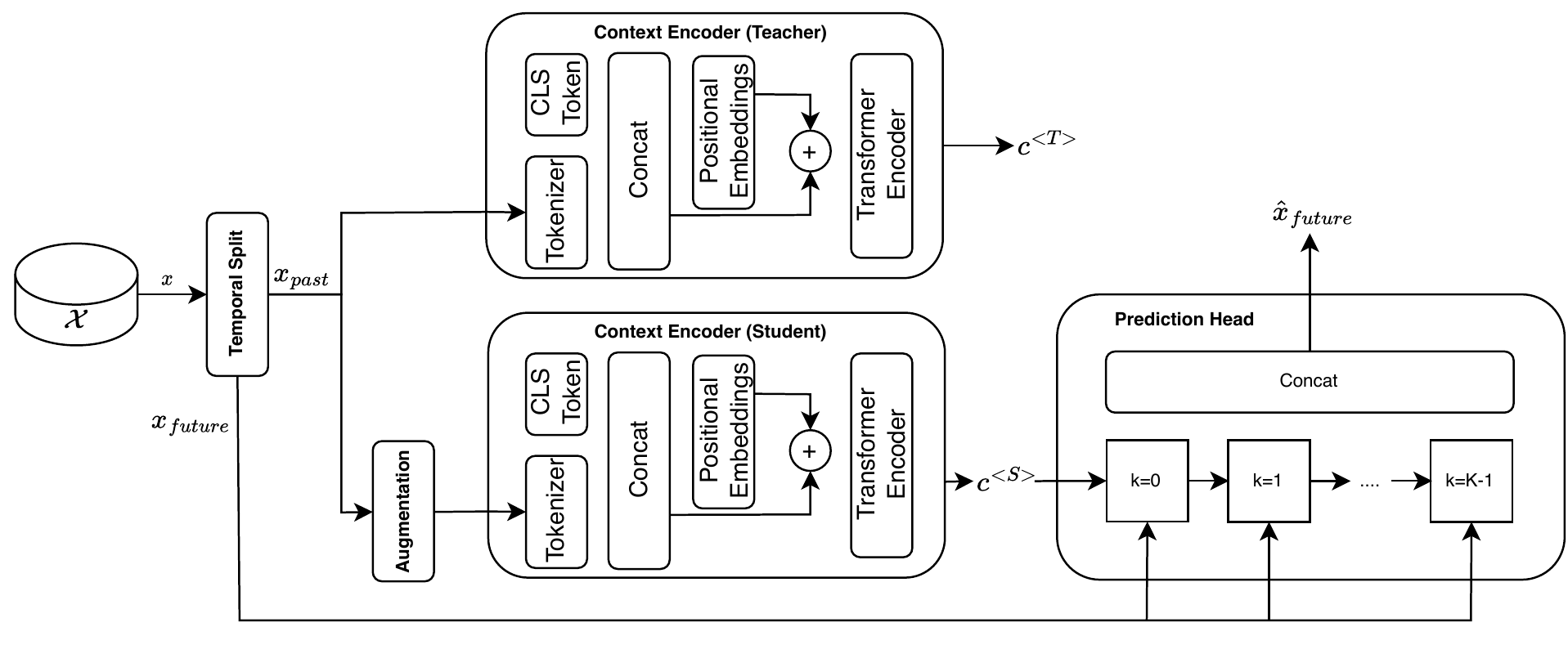}
    \caption{Architecture of the \textit{TS-MoCo} framework consisting of a student and teacher context encoder, and a GRU-based reconstruction head.}
    \label{fig:tsmc-architecture}
\end{figure*}

\section{METHODOLOGY}

\subsection{Training Schema}
We denote the signal with $t$ discrete timesteps by  $x = [x_t]_{t=0}^T$, and signals can be given with labels into $C$ classes.
The training procedure we used is twofold. Firstly, we use a self-supervised pre-training paradigm to learn a generalizable feature encoder. 
Secondly, we optimize a single linear classifier using the representations by this encoder to solve the downstream classification problem.
During self-supervised pre-training, the supervision signal is created directly from the signal itself, omitting the need for laborious annotations. 
We use a combination of two supervision signals.

\subsubsection{Reconstructing the Future}
The first supervision signal is generated by splitting of the final $K$ time-steps from the signal \textit{i.e.} by splitting $x$ into two parts $x_{\text{past}} = [x_t]_{t=0}^{T-K-1}$ and $x_{\text{future}}=[x_t]_{t=T-K}^{T}$. The former is encoded into a representation vector $c$ by a feature encoder described through $f_{\text{ENC}}$ with parameters $\theta$, and the latter is used as target for a reconstruction head, that predicts the subsequent timesteps $t\in[T-K, T]$ of its input from a context vector. 
To compare the predicted subsequent time-steps $\hat{x}_{\text{future}}$ with the ground truth $x_{\text{future}}$ we use a mean squared error loss:
\begin{equation}
    \mathcal{L}_{\text{Rec}} = \lVert x_{\text{future}} - \hat{x}_{\text{future}} \rVert_{2}^2.
\end{equation}

\subsubsection{Momentum Contrast}
The second supervision signal is generated through momentum contrast~\cite{he2020momentum}. 
Two identical architectures, teacher and student models, are used during training.
The teacher model is provided with $x_{\text{past}}$ as input and predicts an output, in our case, a representation vector:
\begin{equation}
    c^{<T>}=f_{\text{ENC}}(x_{\text{past}}:\theta^{<T>}).
\end{equation}
For the student model, the signal is first augmented with an augmentation function before computing the output:
\begin{equation}
c^{<S>}=f_{\text{ENC}}(A(x_{\text{past}}):\theta^{<S>}).    
\end{equation}
Training aims to guide the student towards computing the same representation as the teacher from an augmented version of the input.
We compare these two output vectors using a cosine similarity loss metric:
\begin{equation}
    \mathcal{L}_{\text{MC}} = 1 - cos(c^{<S>}, c^{<T>}).
\end{equation}

\subsubsection{Parameter Update}
After each iteration, student model parameters are updated by loss backpropagation
\begin{equation}
 \mathcal{L}_{\text{SS}}=\mathcal{L}_{\text{Rec}}+\lambda \mathcal{L}_{\text{MC}},   
\end{equation}
where the loss weight $\lambda$ is set as a hyperparameter. The teacher model parameters $\theta^{<T>}$ are updated in order to state an exponentially moving average (EMA) of the student model parameters $\theta^{<S>}$, as utilized in~\cite{he2020momentum}:
\begin{equation}
    \theta^{<T>} = \tau \theta^{<T>} + (1-\tau) \theta^{<S>},
\end{equation}
with $\tau$ being the momentum weight hyperparameter.

\subsubsection{Linear Evaluation}
Following self-supervised pre-training, we employ a linear evaluation scheme to evaluate the learned feature encoder. 
We first use the pre-trained frozen encoder to compute representations of non-augmented input signals $x$. 
Then a single linear layer that predicts class logits from these representations is utilized according to
\begin{equation}
    \hat{y_i} = \log q_i = f_{\text{CLA}}(c^{<S>}_i:\theta^{<C>}),
\end{equation}
where a conventional cross-entropy loss is used.

\subsection{TS-MoCo Framework Architecture}

\subsubsection{Augmentation Function}

Whether or not training the proposed framework is successful heavily depends on the used augmentation function as it defines the encoding and reconstruction task. If the augmentation is too weak, the tasks become trivial and can be solved by the models without learning the underlying concepts of the data. On the other hand, if the augmentation is too strong, training the model may fail due to limited capabilities of the architecture.
Further, we explicitly refrain from domain-specific augmentation functions to allow a fast an easy adaption of our framework to various domains.
By employing a \textit{window-wise temporal masking}, we fulfill aforementioned constraints to the augmentation function. Hereby, a certain percentage $p$ of the signal is overwritten with a masking token \textit{i.e.} $0$. Especially for high-frequency signals, masking singular timesteps scattered along the time-axis would state an easy task, therefore we assure that all masked timesteps span a continuous window.

\subsubsection{Feature Encoder}

The used feature encoder consists of a tokenizer, the addition of positional embeddings, and a transformer encoder. As a first step, the signal used as input to the encoder is mapped into an embedding space by a single linear layer named tokenizer
\begin{equation}
    [h_i^{(1)}]_{t=0}^{T-K-1} = \theta_{\text{tokenizer}} * [x_i]_{t=0}^{T-K-1}.
\end{equation}
Next, like in \cite{devlin2018bert}, a classification token ($<CLS>$) is prepended to the signal. The value of this token is initialized randomly and learned during training.
\begin{equation}
    [h_i^{(2)}] = [\quad<CLS>,\quad [h_i^1]_{t=0}^{T-K-1}\quad]
\end{equation}
In order to allow the feature encoder to exploit positional information of the signal values, we add positional embeddings to the tokenized signals, \textit{i.e.} we add a unique vector to the signal values of each timestep. We create the positional embedding matrix $\mathbf{P}$ from sinusoidal according to previous work~\cite{vaswani2017attention}. We also introduce a hyperparameter $\alpha\in\{0,1\}$, allowing us to enable/disable positional embeddings:
\begin{equation}
    [h_i^{(3)}] = [h_i^{(2)}] + \alpha P_i.
\end{equation}
Finally, the latent signal is input to a transformer encoder \cite{vaswani2017attention} of depth $d$, outputting a contextual embedding vector for each input token $[h_i^{(d+3)}]$. We use the output embedding for the classification token as context encoding $c$.

\subsubsection{Reconstruction Head}
We use Gated Recurrent Units (GRU) for the reconstruction head. The context vector output by the student feature encoder is thereby used as initial hidden state. Further, we apply teacher-forcing for predicting the future timesteps, \textit{i.e.} we pass the ground truth values for a timestep besides a hidden state to the GRU cell for predicting the values of the next timestep.
Our overall architecture is illustrated in~\autoref{fig:tsmc-architecture}.

\begin{table*}[ht!]
\setlength{\tabcolsep}{2pt}
    \centering
    \caption{\textbf{Comparison with Baseline Models.} We compare our self-supervised learning framework with a supervised trained baseline model with the same encoding architecture and a random classifier on the test set.}
    \label{tab:Overview-Results}
    \begin{tabular}{p{2cm} p{3.2cm} p{2cm} c | c | c | c | c}
    \toprule
        \textbf{Dataset} & \textbf{Task} & \textbf{Signal} & \textbf{\# classes / channels} & \textbf{Supervised} & \textbf{TS-TCC} & \textbf{TS-MoCo (ours)} &  \textbf{Random Classifier}\\
    \midrule
        SEED &  Emotion Classification & EEG & 3 / 62 & $0.79$ & $0.42$ & $\mathbf{0.43}$ & $0.33$\\
    UCIHAR  &  Activity Recognition & Inertial Sensory & 6 / 9 & $0.89$ & $0.90$ & $\mathbf{0.52}$ & $0.16$\\ 
    \bottomrule
    \end{tabular}
\end{table*}

\begin{table}[t]
\setlength{\tabcolsep}{5pt}
    \centering
    \caption{\textbf{Ablation Study.} \textit{TS-MoCo} test set accuracies for various settings of relevant hyperparameters, "|" indicates the same value as in the first row.}
    \label{tab:Ablation-Study}
    \begin{tabular}{ccccc|cc}
         \toprule
         \multicolumn{5}{c|}{\textbf{Hyperparameter Settings}} & \multicolumn{2}{c}{\textbf{Classification Accuracies}} \\
         $\kappa$ & $\lambda$ & $K$ & $p_M$ & $\alpha$ & SEED & UCIHAR\\
         \midrule
         $0.9$    & $1.0$     & $6$ & $0.5$ & 0                     & $\mathbf{0.43}$ & $0.36$\\ 
         $0.99$   & |         & |   & |     & |                     & $0.42$          & $0.47$\\ 
         $0.7$    & |         & |   & |     & |                     & $0.42$          & $0.34$\\ 
         |        & $100$     & |   & |     & |                     & $0.41$          & $0.32$\\ 
         |        & $0.01$    & |   & |     & |                     & $0.42$          & $\mathbf{0.52}$\\ 
         |        & |         & $24$& |     & |                     & $0.39$          & $0.37$\\ 
         |        & |         & $2$ & |     & |                     & $0.41$          & $0.37$\\ 
         |        & |         & |   & $0.75$& |                     & $0.41$          & $0.33$\\ 
         |        & |         & |   & $0.25$& |                     & $0.42$          & $0.40$\\ 
         |        & |         & |   & |     & 1                     & $\mathbf{0.43}$ & $0.38$\\ 
         \bottomrule
    \end{tabular}
\end{table}

%% file: sections/03_experimental_study.tex
\section{EXPERIMENTAL STUDY}

\subsection{Datasets}

\subsubsection{SEED}
The SJTU Emotion EEG Dataset (SEED) \cite{zheng2015investigating, duan2013differential} is a dataset of EEG recordings from 15 subjects during viewing of emotion-eliciting videos. Signals were recorded from 62 channels at \SI{1000}{\hertz} and later downsampled to \SI{200}{\hertz}. Following recent work~\cite{bethge2022domain, bethge2022eeg2vec, bethge2022exploiting}, we applied a \SI{4}-\SI{40}{\hertz} bandpass filter and segmented the signals into \SI{2}{\second} non-overlapping windows. EEG recording signals are labeled to be of either negative, neutral, or positive emotions.


\subsubsection{UCIHAR}
This dataset~\cite{anguita2013public} for human activity recognition comprises recordings from 30 subjects during six activities of daily living: \textit{walking, walking upstairs, walking downstairs, sitting, standing, laying}, measured from waist-mounted smartphone accelerometer and gyroscope. Class imbalance is handled via stochastic undersampling~\cite{bethge2022domain}.

\subsection{Baseline Comparison Models}

\subsubsection{Random Classifier}
\label{subsec:dummy-classifier}

We use this as a baseline that is expected to make predictions based on the relative frequency of each class in the training data. 
We employ a stratified strategy, \textit{i.e.} predicted class label is sampled from a multinomial distribution with empirical priors.


\subsubsection{Supervised Trained Models}
\label{subsec:supervised-baselines}

To demonstrate an upper bound of performance, we also compare our framework to a fully-supervised pipeline. 
The architecture of this model is similar to that of the self-supervised trained model during evaluation \textit{i.e.}, it consists of a transformer-based feature encoder followed by a single dense layer for classification.

It is important to note that the supervised, trained model task is much easier to accomplish. However, self-supervised trained models can offer additional advantages of being suitable in scenarios where labeled data is limited and expensive. 
Furthermore, the self-supervised model learns task-agnostic embeddings, which allows the feature encoder to be reused for other tasks arbitrarily. 
We highlight these factors for consideration while interpreting our results.

\subsubsection{Self-Supervised Trained Models}
We compare \textit{TS-MoCo} against another self-supervised learning framework for physiological data, \textit{TS-TCC} as introduced in \autoref{subsec:related-work-phyisological-domain}. In contrast to \textit{TS-MoCo}, \textit{TS-TCC} uses a conventional contrastive loss instead of momentum contrast.

\subsection{Parameter Optimization}
We pre-trained our models for $100$ epochs and then performed $100$ (UCIHAR) / $150$ (SEED) epochs of linear evaluation with early stopping after $20$ epochs based on the validation loss. Baseline models were similarly trained for $100$ epochs. 
We used a $60$-$20$-$20$ split for the train, validation, and test.

%% file: sections/04_results.tex
\section{RESULTS}

\subsection{Comparison with Baseline Models}

\autoref{tab:Overview-Results} presents a comparison of the self-supervised \textit{TS-MoCo} framework with a supervised trained baseline model with same encoding architecture (Supervised), the self-supervised \textit{TS-TCC} framework, and a random classifier based on the strategy introduced in \autoref{subsec:dummy-classifier}. 
Note that the supervised model demonstrates an upper bound of performance to the problems.
Our results clearly show that \textit{TS-MoCo} achieves significantly above chance-level accuracies on both datasets. 
Classification accuracies of the self-supervised models are comparable on the \textit{SEED} dataset, however, the simplified training pipeline of \textit{TS-MoCo} appears to introduce a classification accuracy trade-off on the \textit{UCIHAR} dataset.


\subsection{Ablation Studies}

\subsubsection{Model Components}
To evaluate the influence of the hyperparameters $\kappa$, $\lambda$, $K$, $p_M$, and $\alpha$, we report the results of an ablation study in \autoref{tab:Ablation-Study}. We observe different influences for different domains.
For the \textit{SEED} dataset, varying the hyperparameters barely results in different classification performance, as the best configuration achieves only $+0.04$ better accuracy than the worst configuration.

For \textit{UCIHAR}, we observe results being highly dependent on the choice of the hyperparameters and can result in low accuracies \textit{e.g.}, if the value of $\lambda$ is increased.

\subsubsection{Randomly Initialized Encoder}
To evaluate the feasibility of the pre-training phase of \textit{TS-MoCo}, we compare it against a model with same encoding architecture, but the encoder of this setup is randomly initialized and freezed \textit{i.e.}, only a classifier is fitted to the outputs of a \textit{randomly initialized encoder}. 
By comparing the classification results on the \textit{UCIHAR} dataset, we observe that the classification accuracy of the \textit{random encoder} falls short of that achieved through \textit{TS-MoCo} by $0.06$ ($0.52$ vs $0.46$), \textit{i.e.} pre-training of \textit{TS-MoCo} does help to slightly improve classification accuracy.

%% file: sections/05_discussion.tex
\section{DISCUSSION}
We present \textit{TS-MoCo}\footnote{\url{https://github.com/philipph77/TS-MoCo}}, the first transformer encoder-based self-supervised learning framework with momentum contrast for physiological signal recording domain datasets. 
We performed an experimental pilot study using a linear evaluation scheme to demonstrate the representational capacity of our self-supervised encoding architecture.

We observed that the performance of \textit{TS-MoCo} does not reach to supervised trained baseline models. However importantly, representations learned by \textit{TS-MoCo} offer the additional benefit of being task agnostic and do not require labeled data to be trained.
Further, we note that simplifying the training schema with momentum contrast can limit classification accuracies in certain domains compared to conventional contrastive learning.
We also observed that a strong influence of used hyperparameters can occur depending on the data domain.

Although the supervised baselines performed better than \textit{TS-MoCo}, our model introduces a valuable encoding mechanism for physiological signal domains where no labeled data is available at all when supervised learning is not possible. Especially in the medical domain, labeling data is often cumbersome and dissemination of such information often leads to concerns regarding data privacy~\cite{bencevic2022self}.

In future work, we aim to analyze the amount of labeled data necessary to train a classification layer based on the \textit{TS-MoCo} encoded representations. 
We expect this amount to be significantly lower than needed for the supervised trained models. 
This would favor the use of our pipeline for domains where only very little labeled data is available.